\documentclass[letterpaper,english]{article}
\usepackage[T1]{fontenc}
\usepackage{textcomp}
\usepackage[latin9]{inputenc}
\usepackage{color}
\usepackage{array}
\usepackage{float}
\usepackage{multirow}
\usepackage{varwidth}
\usepackage{amsmath}
\usepackage{amsthm}
\usepackage{amssymb}
\usepackage{graphicx}
\usepackage[numbers]{natbib}
\PassOptionsToPackage{normalem}{ulem}
\usepackage{ulem}

\makeatletter

\pdfpageheight\paperheight
\pdfpagewidth\paperwidth

\providecommand{\tabularnewline}{\\}
\newenvironment{cellvarwidth}[1][t]
    {\begin{varwidth}[#1]{\linewidth}}
    {\@finalstrut\@arstrutbox\end{varwidth}}

\theoremstyle{plain}
\newtheorem{thm}{\protect\theoremname}
\theoremstyle{plain}
\newtheorem{prop}[thm]{\protect\propositionname}

\usepackage{hyperref}
\usepackage{url}
\usepackage{graphicx}

\usepackage{colortbl} 
\definecolor{header_color}{rgb}{0.74,0.88,0.91}
\definecolor{even_color}{rgb}{0.9,0.9,0.9}
\definecolor{subheader_color}{rgb}{0.85,0.93,0.95}
\definecolor{childheader_color}{rgb}{1.0,0.93,0.87}

\ifdefined\showcaptionsetup
 \PassOptionsToPackage{caption=false}{subfig}
\fi
\usepackage{subfig}
\makeatother

\usepackage{babel}
\providecommand{\propositionname}{Proposition}
\providecommand{\theoremname}{Theorem}

\begin{document}
\title{Reward-Guided Survival Data Generation}
\author{Dang Nguyen\textsuperscript{1}, Tu Anh Hoang Nguyen\textsuperscript{1},
Thuc Duy Le\textsuperscript{2}, Sunil Gupta\textsuperscript{1}\\
\textsuperscript{1}Deakin University, Australia\\
\textsuperscript{2}Adelaide University, Australia\textit{}\\
\textit{\{d.nguyen, joseph.nguyen, sunil.gupta\}@deakin.edu.au}\\
\textit{Thuc.Le@Adelaide.edu.au}}
\maketitle
\begin{abstract}
The sharing of tabular survival data is restricted by privacy regulations.
While synthetic data generation offers a promising alternative, existing
methods frequently fail to simultaneously preserve long-term survival
dynamics and patient privacy. We introduce \textbf{SurvTabGen}, a
novel generative framework tailored specifically for survival data
synthesis. Unlike prior multi-stage approaches, SurvTabGen jointly
models covariates, event indicators, and survival time within a unified
structurally aware generator. We first extract causal relationships
among covariates via causal discovery. The generation then follows
a dependency-aware sequence: event outcomes are conditioned on covariates,
and survival times are conditioned on both covariates and events.
Crucially, the generator is trained with a reinforcement learning-based
survival-aware objective. By using the Integrated Absolute Difference
(IAD) between real and synthetic survival curves as a reward signal,
the model directly optimizes for macroscopic temporal fidelity. Extensive
evaluations against 10 SOTA baselines across seven diverse real-world
datasets demonstrate the superiority of SurvTabGen. Our results demonstrate
that SurvTabGen provides a highly effective solution for generating
high-fidelity, privacy-preserving synthetic cohorts that excel in
survival preservation and data quality.
\end{abstract}

\section{Introduction\label{sec:Introduction}}

\begin{figure}[th]
\begin{centering}
\subfloat[Event and censoring]{\begin{centering}
\includegraphics[width=0.35\columnwidth]{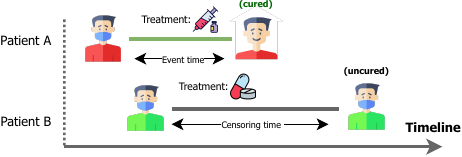}
\par\end{centering}
}\subfloat[Over-optimistic]{\begin{centering}
\includegraphics[width=0.2\columnwidth]{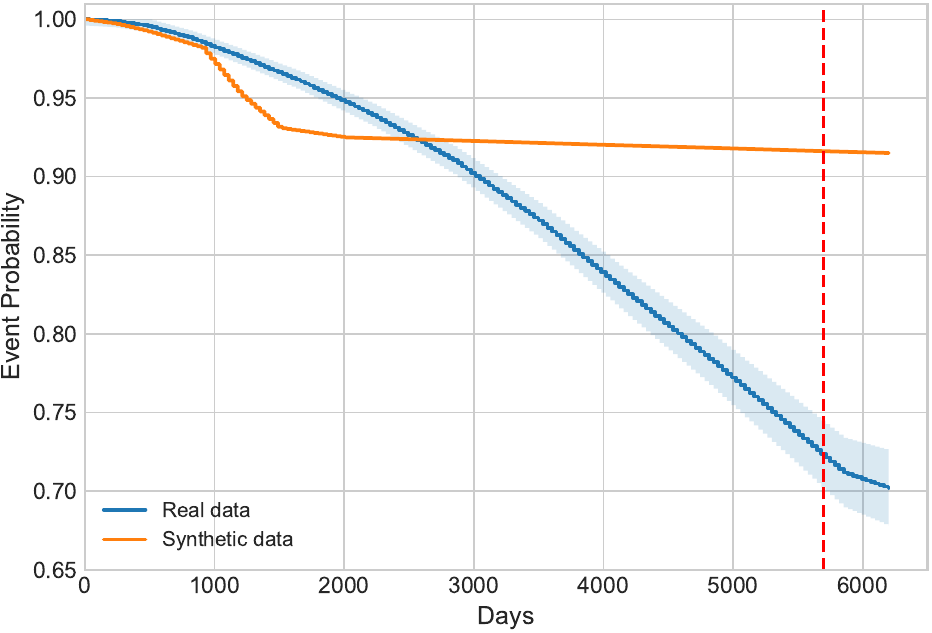}
\par\end{centering}
}\subfloat[Over-pessimistic]{\begin{centering}
\includegraphics[width=0.2\columnwidth]{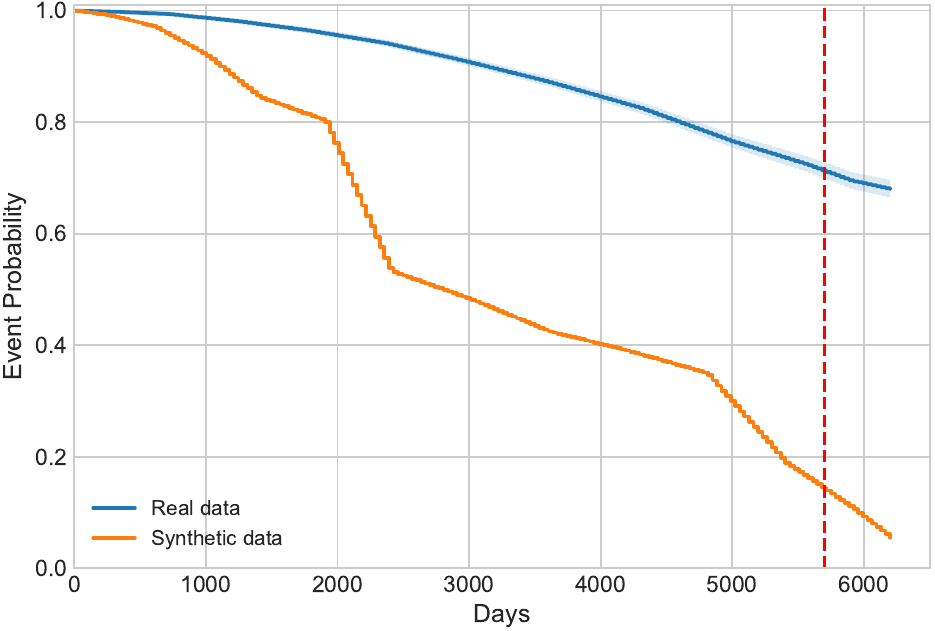}
\par\end{centering}
}\subfloat[Short-sightedness]{\begin{centering}
\includegraphics[width=0.2\columnwidth]{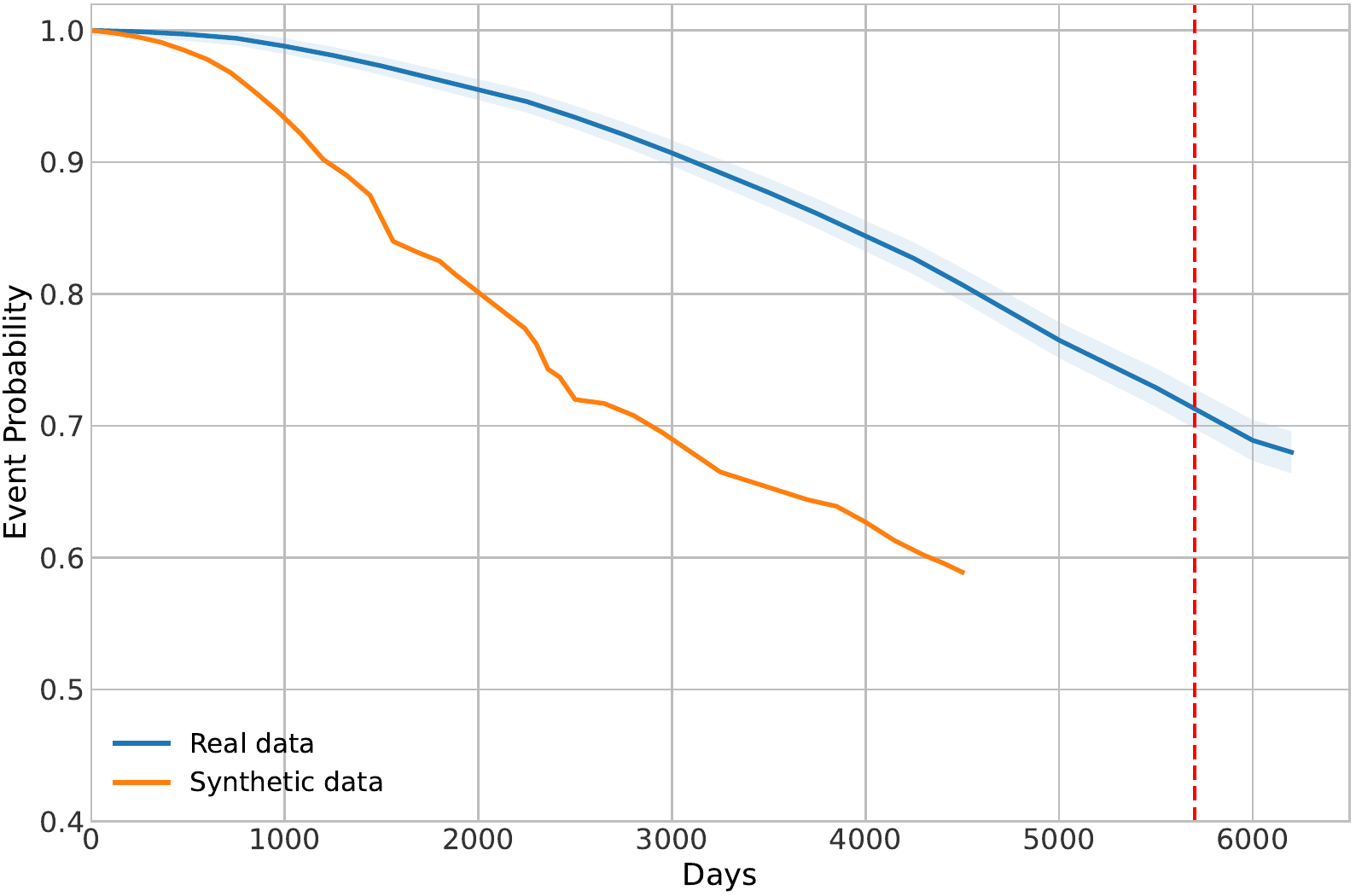}
\par\end{centering}
}
\par\end{centering}
\caption{\label{fig:Illustration-of-temporal}\textbf{(a)} Illustration of
the competing event and censoring mechanisms in survival data. \textbf{(b-d)}
Common failure modes of temporal bias in synthetic survival data.}
\end{figure}

\textit{Survival data} capture the time until a predefined event occurs,
frequently accompanied by \textit{censoring} due to incomplete observations
\citep{hahn2025generating}. This data consists of descriptive features
(e.g., demographic, treatments, lab results), collectively termed
\textit{covariates} $X$, alongside two critical features: the \textit{event
indicator} $E$ and the \textit{time} $T$. When an event observed
(i.e., $E=1$, such as a patient recovering from a disease), $T=t$
denotes the exact event time. Conversely, when the event is not observed
within the study period (i.e., $E=0$), $T=t$ becomes censoring time,
indicating that the patient survived without experiencing the event
up to time $t$, as illustrated in Figure \ref{fig:Illustration-of-temporal}(a).
Survival data are foundational to research in healthcare, epidemiology,
and reliability engineering. However, these datasets are costly and
time-consuming to collect. More critically, they often contain highly
sensitive personal information, limiting their availability and restricting
open data sharing \citep{vallevik2024can}. To overcome these barriers,
synthetic data generation has emerged as a solution, enabling the
creation of realistic, privacy-preserving survival data that can serve
as viable proxies for restricted real-world datasets \citep{norcliffe2023survivalgan}.

A primary application of such data is \textit{survival analysis}.
Given a survival dataset, this analysis seeks to answer fundamental
prognostic questions, such as: ``\textit{What is the probability
a patient will survive past time $t$?}''. This insight is critical
for drug development, treatment efficacy estimation, and risk assessment.
Consequently, any synthetic survival data must accurately preserve
these macroscopic survival dynamics. However, generating high-fidelity
survival data remains a profound challenge due to the presence of
censoring. Standard tabular generative models such as CTGAN \citep{Xu2019},
TabDiff \citep{shi2025tabdiff}, and GReaT \citep{Borisov2023} often
fail in this domain because they treat every column as a definitive,
fully observed ground truth. Furthermore, real-world survival data
exhibits a severe structural imbalance between observed events and
censored samples \citep{eckardt2024mimicking}. If a generative model
fails to capture this imbalance, it introduces severe temporal bias
into the synthetic data, as evidenced by deviations in the Kaplan-Meier
survival curves between real and synthetic cohorts (Figure \ref{fig:Illustration-of-temporal}(b-d)).
This failure typically manifests in three distinct ways: \textbf{(1)
Over-optimism}: Synthetic events occur later than in reality, inflating
the predicted expected lifetime, \textbf{(2) Over-pessimism:} Synthetic
events occur earlier than expected, deflating the survival probability,
and \textbf{(3) Short-sightedness:} The model fails to generate long-term
horizons, prematurely collapsing the survival curve. These temporal
distortions complicate the learning of accurate time-to-event distributions
and mislead clinical applications.

To the best of our knowledge, only three models SurvivalGAN \citep{norcliffe2023survivalgan},
the Ashhad framework \citep{ashhad2025generating}, and SurvDiff \citep{brockschmidt2026survdiff}
have been proposed for survival data generation. To mitigate temporal
distortions, both SurvivalGAN and Ashhad rely on empirical pre-sampling.
By drawing the event indicator $E$ (and the time $T$ in Ashhad's
case) directly from the real data's empirical distribution, these
models force the synthetic dataset to inherit the real-world censoring
ratio. However, this approach introduces two critical vulnerabilities.
First, anchoring the generation process to the empirical distribution
increases the risk of data memorization, compromising the privacy-preserving
mandate of synthetic data generation. Second, this forces an unnatural
generative sequence. Sampling outcomes $E$ first and generating the
covariates $X$ conditioned on those outcomes requires the model to
learn the \textit{reverse-conditional distribution} $p(X\mid E)$.
This violates the natural chronological progression of clinical reality
(where baseline health dictates future outcomes) and is structurally
more difficult to optimize than the forward prospective distribution
$p(E\mid X)$. Alternatively, SurvDiff samples from the joint distribution
$p(X,E,T)$ via a diffusion process. However, it treats all features
as structurally independent and lacks a mechanism to explicitly minimize
macroscopic survival curve deviations, leaving it vulnerable to the
temporal biases illustrated in Figure \ref{fig:Illustration-of-temporal}(b-d).

To resolve these limitations, we propose Survival Tabular Data Generator
(\textbf{SurvTabGen}), an end-to-end framework that models survival
data within a unified generative process. Unlike prior approaches
that rely on externally sampled event indicators or reverse-conditional
generation, our method enforces the causal dependency structure of
clinical data. Specifically, covariates are generated first via structured
sub-generators informed by a causal graph (DAG). Event outcomes are
then generated conditionally on these covariates, and survival times
are generated conditionally on both the covariates and the event outcomes.
This sequential DAG architecture ensures that structural dependencies
are respected. Furthermore, we introduce a survival-aware training
objective that explicitly aligns population-level survival behavior
of the real and synthetic cohorts by minimizing the Integrated Absolute
Difference (IAD) between their Kaplan-Meier curves. Because this objective
is non-differentiable, we optimize it via policy gradients, treating
survival curve discrepancy as a global Reinforcement Learning (RL)
reward. This formulation enables the model to directly optimize macroscopic
temporal fidelity rather than relying solely on marginal distribution
matching.

We summarize \textbf{our contributions} as follows:

1. \textbf{Architectural innovation:} We propose SurvTabGen, an end-to-end
generative framework that explicitly extracts and preserves structural
causal dependencies for survival data generation.

2. \textbf{Reward-guided optimization:} We develop a novel survival-aware
training objective that aligns survival behavior between real and
synthetic cohorts using non-differentiable metrics (IAD), optimized
via policy gradients.

3. \textbf{Empirical validation:} We conduct extensive experiments
across seven diverse real-world datasets, demonstrating that SurvTabGen
is better than existing methods in survival fidelity.

4. \textbf{Novelty:} Compared to SurvivalGAN and Ashhad, SurvTabGen
trains a single, unified model to learn the joint distribution $p(X,E,T)$,
avoiding overly complex multi-stage or reverse-conditional frameworks.
Compared to SurvDiff, SurvTabGen respects the causal relationship
among features and explicitly minimize macroscopic curve deviations
to neutralize the temporal biases of over-optimism, over-pessimism,
and short-sightedness.

\section{Related Works\label{sec:Related-Works}}

\paragraph{Tabular data generative models}

Generative models for tabular data synthesis include GAN/VAE architectures:
CTGAN \citep{Xu2019}, CA-GAN \citep{nguyen2026causal}, and TVAE
\citep{Xu2019}, diffusion frameworks: TabDiff \citep{shi2025tabdiff},
and LLMs: GReaT \citep{Borisov2023}, TapTap \citep{zhang2023generative},
and Pred-LLM \citep{nguyen2024tabular}. While these general-purpose
models can reproduce marginal distributions \citep{eckardt2024mimicking},
they struggle with long-term censoring dynamics and complex time-to-event
trajectories. Because they rely on generic reconstruction rather than
explicitly modeling the joint structure of covariates, event indicators,
and survival times \citep{hahn2025generating,ahmed2026identifying},
they frequently violate temporal constraints and fail to achieve true
survival consistency \citep{vallevik2024can}.\vspace{-0.2cm}

\paragraph{Survival data generative models}

To address censoring, specialized frameworks like SurvivalGAN \citep{norcliffe2023survivalgan}
and the Ashhad method \citep{ashhad2025generating} use multi-stage
pipelines that model covariates and temporal outcomes as disjoint
components. Both rely on the empirical pre-sampling of event indicators,
which propagates compounding errors across disconnected sub-models
and elevates privacy leakage risks by anchoring synthetic data to
the original cohort. Alternatively, SurvDiff \citep{brockschmidt2026survdiff}
employs end-to-end diffusion to jointly generate all variables from
noise. However, SurvDiff treats all features as structurally independent,
ignoring causal dependencies, and lacks a mechanism to explicitly
enforce macroscopic temporal consistency. Consequently, current SOTA
methods remain largely distribution-driven; they fail to simultaneously
enforce the structural causal relationships and the population-level
temporal alignment necessary for high-fidelity survival data generation.

\section{Framework\label{sec:Framework}}

\subsection{Problem Formulation}

Given a tabular survival dataset $\mathcal{D}=\{u_{i}\}_{i=1}^{N}$,
where each instance $u_{i}=(x_{i},t_{i},e_{i})$ is drawn from an
unknown data distribution $p_{\mathrm{data}}$, the primary objective
is to learn a generative model that produces synthetic samples $\hat{\mathcal{D}}=\{\hat{u}_{i}\}_{i=1}^{N}$
following the same joint distribution. Unlike prior works \citep{norcliffe2023survivalgan,ashhad2025generating,brockschmidt2026survdiff},
we assume the underlying data is governed by a causal structure over
$(X,T,E)$, inducing dependencies that extend beyond simple statistical
correlations. Accordingly, our goal is to learn a generator $G_{\theta}$
that preserves not only marginal distributions but also directional
dependencies linking covariates with event occurrence and event time.
Crucially, this requires capturing the complex survival dynamics where
$X$, $E$, and $T$ jointly determine the survival behavior. This
entails aligning macroscopic survival patterns of the synthetic data
with the real data, moving beyond conventional tabular distribution
matching.

\subsection{Proposed Method: Survival Tabular Data Generator (SurvTabGen)}

\begin{figure}[th]
\begin{centering}
\includegraphics[width=0.6\columnwidth]{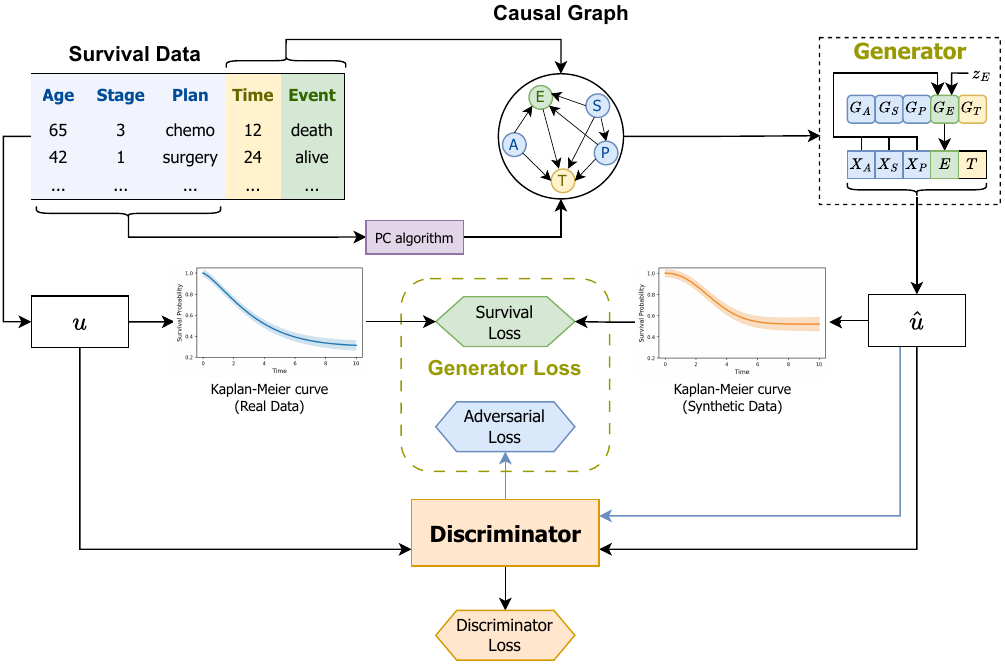}
\par\end{centering}
\caption{\label{fig:Framework}Overview of SurvTabGen. \textbf{(1) Causal Discovery:}
A directed acyclic graph (DAG) is extracted from real survival covariates
to map the underlying structural dependencies. \textbf{(2) Structured
Generation:} A Wasserstein GAN produces covariates, event indicators,
and survival times autoregressively along the DAG while a discriminator
enforces distributional realism. \textbf{(3) Survival Alignment:}
A reinforcement learning-based survival loss translates Kaplan-Meier
curve discrepancies into batch-level rewards.}
\end{figure}

We propose \textbf{SurvTabGen}, an end-to-end generative framework
that integrates structured dependency modeling with a survival-aware
optimization loop. Given input noise, the model jointly generates
covariates, event indicators, and survival time through a unified,
causally consistent generative process. To enforce realistic survival
behavior, SurvTabGen leverages a reinforcement learning strategy,
utilizing survival curve discrepancies as reward signals to guide
training. The overview of the framework is illustrated in Figure \ref{fig:Framework}.

\subsubsection{Structural dependency modeling}

A foundational component of SurvTabGen is the dependency structure
governing how variables are generated. While one could infer a single
dependency graph over all variables, applying standard causal discovery
algorithms (e.g., the Peter-Clark (PC) algorithm \citep{spirtes2000causation})
directly to the joint space of $(X,E,T)$ may be problematic. In survival
data, $T$ and $E$ are coupled by the censoring mechanism, meaning
the observed $T$ represents a mixture of true event times and censoring
times. This deterministic masking violates the standard conditional
independence assumptions (e.g., faithfulness) required by PC, frequently
resulting in spurious edges or inverted causal directions between
baseline features and clinical outcomes. Furthermore, the chronological
flow from covariates ($X$) to temporal outcomes ($E$ and $T$) is
an established clinical reality. Rather than forcing an algorithm
to blindly rediscover this chronology from noisy, censored data, it
is more robust to enforce it as a structural inductive bias. Therefore,
we restrict the data-driven causal discovery phase entirely to the
covariate space. Let $X=(X_{1},\dots,X_{M})$ denote the feature variables.
We apply PC to $X$ to infer a directed acyclic graph (DAG) $\mathcal{G}_{X}=(\mathcal{V}_{X},\mathcal{E}_{X})$.
From this, we obtain the parent sets $\mathrm{PA}_{i}$ for each $X_{i}$
and establish a valid topological ordering to guide the generation
phase.

While $\mathcal{G}_{X}$ captures dependencies among covariates, it
does not specify the generation sequence for survival variables. Instead
of leaving these relations implicit, we encode them directly using
a survival-oriented inductive bias: the event indicator depends on
the covariates, and the time-to-event depends on both the covariates
and the event indicator. Formally, we augment $\mathcal{G}_{X}$ with
the edges $X\rightarrow E$ and $(X,E)\rightarrow T$, yielding the
final graph $\mathcal{G}$ with $\mathrm{PA}_{E}=X$ and $\mathrm{PA}_{T}=X\cup\{E\}$.
This provides a logical dependency backbone for the generation phase.

\subsubsection{Wasserstein GAN (WGAN) architecture}

SurvTabGen adopts a modular adversarial architecture inspired by CA-GAN
\citep{nguyen2026causal}. It consists of a discriminator $D$ alongside
feature-specific sub-generators $\{G_{i}\}_{i=1}^{M}$ for covariates,
and two terminal sub-generators $G_{E}$ and $G_{T}$ for the event
indicator and time-to-event, respectively. Let $\pi$ denote a valid
topological ordering of the causal graph $\mathcal{G}$. For each
feature $X_{i}$, the corresponding sub-generator $G_{i}$ models
the conditional generation of $X_{i}$ given $\mathrm{PA}_{i}$, utilizing
the generated values of its parent variables and an independent noise
vector $z_{i}\sim\mathcal{N}(0,I)$. The feature and survival variables
are generated as follows:
\begin{equation}
\hat{X}_{i}=G_{i}(\mathrm{PA}_{i},z_{i}),\text{ }\hat{E}=G_{E}(\hat{X}_{1},\dots,\hat{X}_{M},z_{E}),\text{ }\hat{T}=G_{T}(\hat{X}_{1},\dots,\hat{X}_{M},\hat{E},z_{T}),\label{eq:Generate_X}
\end{equation}
where $z_{i,}z_{E},z_{T}\sim\mathcal{N}(0,I)$. The full synthetic
sample $\hat{u}=(\hat{X}_{1},\dots,\hat{X}_{M},\hat{E},\hat{T})$
is constructed autoregressively along $\pi$, ensuring structural
consistency with the dependencies encoded in $\mathcal{G}$.

\subsubsection{Training the WGAN}

\paragraph{Training discriminator}

We train $D$ using the Wasserstein GAN with Gradient Penalty (WGAN-GP)
objective \citep{Gulrajani2017}. The discriminator is optimized by
evaluating real and generated samples via:
\begin{equation}
\mathcal{L}_{D}=-\mathbb{E}_{u\sim p_{\mathrm{data}}}[D(u)]+\mathbb{E}_{\hat{u}\sim p_{\theta}}[D(\hat{u})]+\mathcal{L}_{\mathrm{gp}},\label{eq:Discriminator_loss}
\end{equation}
where $\mathcal{L}_{\mathrm{gp}}$ denotes the standard gradient penalty
defined on interpolated samples. This objective provides a stable
adversarial signal for training the structured generator.

\paragraph{Training generator}

Synthetic samples $\hat{u}=(\hat{X},\hat{E},\hat{T})$ are generated
following the topological ordering $\pi$ of $\mathcal{G}$. These
samples update the generator through the adversarial objective, which
encourages higher critic scores to reduce the discrepancy between
the model distribution $p_{\theta}$ and the real data distribution
$p_{\mathrm{data}}$:
\begin{equation}
\mathcal{L}_{\mathrm{adv}}=-\mathbb{E}_{\hat{u}\sim p_{\theta}}[D(\hat{u})],\label{eq:Adv_loss}
\end{equation}
Following the WGAN-GP training scheme, the discriminator is updated
multiple times per generator step to ensure stable gradients.

\subsubsection{Survival-preserving objective for the generator}

Standard Wasserstein matching is often insufficient to capture macroscopic
temporal dynamics under censoring. To mitigate this, we extend the
generator objective with a survival-aware regularization term:
\begin{equation}
\mathcal{L}_{G}=\mathcal{L}_{\mathrm{adv}}+\lambda\mathcal{L}_{\mathrm{surv}},\label{eq:Generator_loss}
\end{equation}
where $\mathcal{L}_{\mathrm{surv}}$ quantifies the discrepancy in
survival behavior between the real and generated cohorts. The coefficient
$\lambda$ determines the strength of this constraint, allowing the
model to balance feature-level distributional realism with macroscopic
temporal accuracy.

Previous works modeled individual-level conditional survival functions
using separate components such as DeepHit \citep{norcliffe2023survivalgan}
or one-dimensional generators \citep{ashhad2025generating}. These
multi-stage designs prevent survival behavior from being directly
optimized within the generator, risking misalignment with the generated
features.

In contrast, we leverage the population-level marginal survival function
$S(t)=P(T>t)$ as the direct alignment target. Although defined marginally,
this function reflects the complex temporal behavior induced by the
underlying joint distribution over $(X,E,T)$ because both the event
occurrence and the time-to-event are generated conditionally on the
covariates through the estimated DAG structure. Aligning $S(t)$ therefore
provides a stable and low-variance signal under censoring while encouraging
consistency with the temporal behavior induced by the joint distribution.

We estimate $S(t)$ with the Kaplan-Meier estimator, using observed
time-event pairs $\{(t_{i},e_{i})\}_{i=1}^{N}$:
\begin{equation}
S(t)=\prod_{t_{i}\le t}\left(1-\frac{d_{i}}{n_{i}}\right),\label{eq:KM_estimator}
\end{equation}
where $d_{i}$ denotes the number of events at time $t_{i}$ and $n_{i}$
is the number at risk just prior to $t_{i}$. The estimator is evaluated
on a discrete grid $\mathcal{T}=\{t_{1},\dots,t_{K}\}$, constructed
from either sorted event times or a uniform partition of the observed
range to ensure consistent comparison between datasets.

The discrepancy between survival curves is quantified using \textit{integrated
absolute difference} (IAD):
\begin{equation}
\mathrm{IAD}=\int\left|S_{\mathrm{real}}(t)-S_{\mathrm{syn}}(t)\right|dt\label{eq:IAD}
\end{equation}
which is approximated numerically on the discrete grid $\mathcal{T}$
using the trapezoidal rule. This formulation measures global deviations
in temporal dynamics while remaining sensitive to differences in tail
behavior and censoring patterns.

To theoretically justify the use of IAD as a training signal, we establish
a formal connection between this metric and the optimal transport
framework of our underlying WGAN.
\begin{prop}
\label{prop:wasserstein_iad}Let $P_{T}$ and $Q_{T}$ denote the
true and synthetic marginal distributions of the event time $T$ (both
with finite first moments), with respective survival functions $S_{\mathrm{real}}(t)$
and $S_{\mathrm{syn}}(t)$. The continuous form of the IAD coincides
with the 1-Wasserstein distance between the corresponding one-dimensional
time-to-event distributions: $\mathrm{IAD}=\mathcal{W}_{1}(P_{T},Q_{T})$.
\end{prop}

In practice, when survival times are subject to censoring, we approximate
$S_{\mathrm{real}}(t)$ and $S_{\mathrm{syn}}(t)$ using Kaplan-Meier
estimators, allowing the empirical IAD to serve as a proxy for the
Wasserstein distance between the underlying (latent) time-to-event
distributions.

Proposition \ref{prop:wasserstein_iad} (\textit{proof in Appendix
\ref{sec:Proof-for-Theorem}}) reveals a theoretical insight: while
the discriminator minimizes the \textit{joint} high-dimensional distance,
the curse of dimensionality often prevents WGANs from perfectly matching
specific margins under censoring. By utilizing this Wasserstein-aligned
temporal metric (IAD), we construct a targeted reward signal that
directly optimizes this temporal margin.

Because the Kaplan-Meier estimator requires a population to compute,
we define a shared \textbf{batch-level reward} for a generated batch
$\hat{\mathcal{B}}=\{\hat{u}_{1},\dots,\hat{u}_{B}\}$ as: $R(\hat{\mathcal{B}})=-\mathrm{IAD}$.
This shared reward is uniformly assigned to all samples within the
batch. Our \textbf{survival loss} is therefore defined over the batch
as: $\mathcal{L}_{\mathrm{surv}}=-\mathbb{E}_{\hat{\mathcal{B}}\sim p_{\theta}}\left[R(\hat{\mathcal{B}})\right]$.

However, the survival loss cannot be optimized via standard backpropagation
because the reward depends on the \textit{non-differentiable} Kaplan-Meier
procedure and the IAD metric. To optimize $G$, we cast the generation
process into a reinforcement learning (RL) paradigm. Because standard
WGAN generators are implicit models, the true generative likelihood
$p_{\theta}(\hat{u})$ required for standard policy gradients is intractable.
We resolve this by interpreting the terminal layers of the generator
as a stochastic policy, thereby constructing a tractable \textit{surrogate
likelihood}. Specifically, for \textbf{categorical variables}, we
employ the Gumbel-Softmax trick \citep{jang2016categorical}. For
\textbf{continuous variables}, the generator outputs parameterize
a stochastic distribution (e.g., Gaussian). While this formulation
acts as an approximation rather than the true model likelihood, it
provides a valid surrogate log-probability for a generated sample
$\hat{u}$. This enables optimization via the score-function estimator:
\begin{equation}
\log p_{\theta}(\hat{u})=\sum_{i=1}^{M}\log p_{i}(\hat{X}_{i}\mid\mathrm{PA}_{i})+\log p_{E}(\hat{E}\mid\mathrm{PA}_{E})+\log p_{T}(\hat{T}\mid\mathrm{PA}_{T})\label{eq:log_p}
\end{equation}

To stabilize the policy gradient, we augment the reward signal with
an auxiliary penalty term $\mathcal{P}(\hat{\mathcal{B}})$ that jointly
regularizes discrepancies in marginal event rates and conditional
time distributions. Formal definition of $\mathcal{P}(\hat{\mathcal{B}})$
is provided in \textit{Appendix} \ref{sec:aux_penalty}. We optimize
the generator using the REINFORCE policy gradient algorithm \citep{Williams1992}:
\begin{equation}
\nabla_{\theta}\mathcal{L}_{\mathrm{surv}}=-\mathbb{E}_{\hat{\mathcal{B}}\sim p_{\theta}}\left[(R(\hat{\mathcal{B}})-\mathcal{P}(\hat{\mathcal{B}}))\frac{1}{B}\sum_{b=1}^{B}\nabla_{\theta}\log p_{\theta}(\hat{u}_{b})\right]\label{eq:final_survival_loss}
\end{equation}

Overall, SurvTabGen integrates structural generation with a survival-aware
RL optimization loop. By combining adversarial training with a reward-driven
objective, it captures both underlying distribution and macroscopic
temporal dynamics. Consequently, by minimizing the discrepancy between
the survival curves, it helps mitigate the three failure modes in
synthetic survival data generation: \textit{over-optimism}, \textit{over-pessimism},
and \textit{short-sightedness}, as illustrated previously in Figure
\ref{fig:Illustration-of-temporal}(b-d).

\section{Experiments\label{sec:Experiments}}

\subsection{Experiment Settings}

\paragraph{Datasets}

We evaluate SurvTabGen on seven datasets, including \textit{AIDS}
\citep{hosmer2008applied}, Chronic Graft-versus-Host Disease (\textit{Cgvhd})
\citep{pintilie2006competing}, Serum Free Light Chain (\textit{Flchain})
\citep{dispenzieri2012use}, German Breast Cancer Study Group 2 (\textit{Gbsg2})
\citep{schumacher1994randomized}, Veterans\textquoteright{} Lung
Cancer (\textit{Veterans}) \citep{kalbfleisch2002statistical}, Worcester
Heart Attack Study (\textit{Whas500}) \citep{hosmer2008applied},
and Additively Manufactured (\textit{AM}) \citep{zhang2023fatigue}.
All datasets contain mixed-type features (continuous and categorical)
and are often used in prior synthetic survival literature \citep{norcliffe2023survivalgan,ashhad2025generating,brockschmidt2026survdiff}.
Their full descriptions are provided in \textit{Appendix} \ref{sec:Data-descriptions}.
Notably, \textit{while prior methods were tested on only 3-5 datasets
strictly within the healthcare domain, SurvTabGen is rigorously evaluated
across 7 datasets spanning both healthcare and engineering applications}.

\paragraph{Evaluation metrics}

Following previous works \citep{norcliffe2023survivalgan,ashhad2025generating,brockschmidt2026survdiff},
we evaluate the generated data across three aspects. \textbf{Survival
fidelity:} Assessed using Kaplan-Meier based metrics \citep{norcliffe2023survivalgan}
such as \textit{KM Divergence}, \textit{Optimism}, and \textit{Short-sightedness},
where values closer to zero indicate superior survival preservation.\textbf{
Data quality:} Evaluated in terms of distributional fidelity using
\textit{Jensen-Shannon} (JS) and \textit{Wasserstein} distances, where
lower values are optimal. \textbf{Privacy:} Assessed using the \textit{Distance
to Closest Record} (DCR) \citep{yang2025doubling} and \textit{Re-identification
risk} \citep{Yoon2020}. While excessively low DCR values may indicate
potential privacy leakage, excessively high values suggest poor realism.
Re-identification risk evaluates whether synthetic samples can be
linked back to real individuals. Detailed formulations for these metrics
are provided in \textit{Appendix} \ref{sec:Metrics}.

\paragraph{Baselines}

We benchmark SurvTabGen against 10 recent SOTA baselines spanning
four generative paradigms: \textbf{(1) VAE and GAN-based models:}
TVAE \citep{Xu2019}, CTGAN \citep{Xu2019}, and CA-GAN \citep{nguyen2026causal},
\textbf{(2) LLM-based approaches:} GReaT \citep{Borisov2023}, TapTap
\citep{zhang2023generative}, and Pred-LLM \citep{nguyen2024tabular},
\textbf{(3) Diffusion-based methods:} TabDiff \citep{shi2025tabdiff},
and \textbf{(4) Survival-aware methods:} SurvivalGAN \citep{norcliffe2023survivalgan},
the Ashhad framework \citep{ashhad2025generating}, and SurvDiff \citep{brockschmidt2026survdiff}.
SurvivalGAN serves as our main competitor as it also uses a GAN backbone.

\paragraph{Implementation details}

The discriminator is a 3-layer MLP with two hidden layers of 256 units
while each sub-generator is a 4-layer MLP. For output activations,
we apply \textit{tanh} for continuous variables and \textit{Gumbel-Softmax}
trick for categorical variables. The model is trained for 500 epochs
with a noise dimension of 16 and a batch size of 500. The survival
coefficient is fixed at $\lambda=0.1$ (sensitivity analysis is provided
in \textit{Appendix} \ref{subsec:Different-lambda}). We run PC with
$\alpha=0.05$ to extract the estimated DAG from the covariates (alternative
causal discovery methods are investigated in \textit{Appendix} \ref{subsec:Different-causal-discovery}).
We repeat each experiment 5 times and report the average score along
with standard deviation.

\subsection{Results and Discussions}

\paragraph{Survival preservation}

As shown in Table \ref{tab:KM-Divergence-scores}, SurvTabGen achieves
the lowest average KM Divergence score, indicating better alignment
with real-world survival distributions. The Ashhad method performs
well, securing the second-best average. Because Ashhad samples $E$
and $T$ from empirical distributions, it may copy the time-event
distribution of the real dataset. Therefore, it introduces a critical
weakness: the resulting synthetic datasets map too closely to the
real datasets, increasing information leakage. Our subsequent privacy
evaluation highlights this vulnerability.

\begin{figure}[th]
\begin{centering}
\includegraphics[width=1\textwidth]{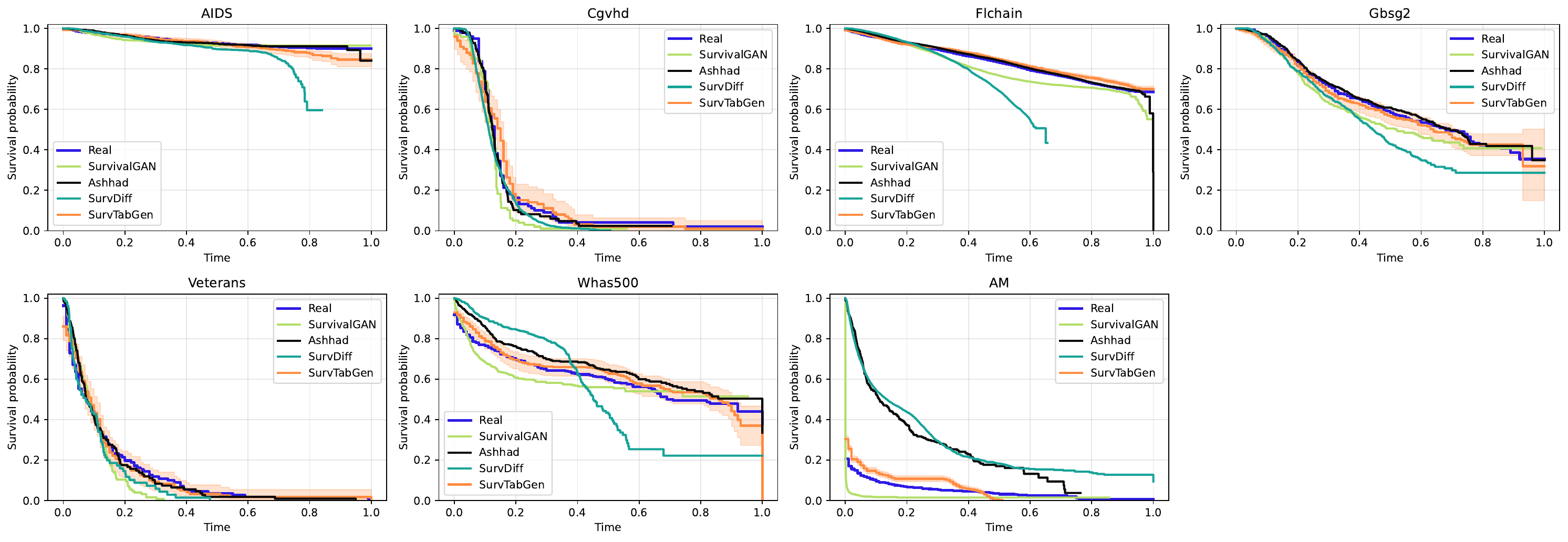}
\par\end{centering}
\caption{\label{fig:Kaplan=002013Meier}Kaplan-Meier curves across all datasets.
SurvTabGen closely preserves the real survival trajectories whereas
other methods exhibits deviations. Time is scaled to $[0,1]$ for
consistent representation.}
\end{figure}

Overall, SurvTabGen outperforms SurvivalGAN by a 3\% margin. This
advantage is demonstrated in Figure \ref{fig:Kaplan=002013Meier}.
While SurvTabGen accurately tracks the real Kaplan-Meier curves across
diverse datasets, capturing both early-onset dynamics and long-term
tail behavior, SurvivalGAN exhibits substantial deviations, particularly
in later stages. These results suggest that incorporating causal dependencies
and a survival-aware objective yields temporal fidelity superior to
standard distribution matching.

\begin{table}[th]
\caption{\label{tab:KM-Divergence-scores}KM Divergence scores ($\downarrow$)
showing the survival preservation performance of each method. \textbf{Bold}
and \uline{underline} indicate the best and second-best methods.}

\centering{}\resizebox{\textwidth}{!}{%
\begin{tabular}{|l|c|c|c|c|c|c|c|c|}
\hline 
\rowcolor{header_color} & AIDS & Cgvhd & Flchain & Gbsg2 & Veterans & Whas500 & AM & \textcolor{red}{AVG}\tabularnewline
\hline 
\hline 
CTGAN & 0.046 (0.01) & 0.044 (0.02) & 0.079 (0.02) & 0.087 (0.02) & 0.091 (0.07) & 0.133 (0.04) & 0.104 (0.01) & \textcolor{red}{0.084 (0.03)}\tabularnewline
\rowcolor{even_color}CA-GAN & 0.148 (0.01) & 0.273 (0.18) & 0.099 (0.01) & 0.185 (0.06) & 0.307 (0.07) & 0.104 (0.07) & 0.096 (0.01) & \textcolor{red}{0.173 (0.06)}\tabularnewline
TVAE & 0.055 (0.01) & 0.053 (0.00) & 0.057 (0.01) & 0.075 (0.01) & 0.078 (0.00) & \uline{0.040} (0.01) & 0.061 (0.01) & \textcolor{red}{0.060 (0.01)}\tabularnewline
\rowcolor{even_color}TabDiff & 0.159 (0.06) & 0.140 (0.11) & 0.086 (0.01) & 0.231 (0.08) & 0.265 (0.10) & 0.147 (0.04) & 0.283 (0.06) & \textcolor{red}{0.187 (0.07)}\tabularnewline
GReaT & 0.059 (0.00) & 0.034 (0.00) & 0.026 (0.00) & 0.109 (0.00) & 0.043 (0.00) & 0.102 (0.00) & \textbf{0.012} (0.00) & \textcolor{red}{0.055 (0.00)}\tabularnewline
\rowcolor{even_color}TapTap & 0.065 (0.00) & 0.142 (0.00) & 0.174 (0.00) & 0.257 (0.00) & 0.100 (0.00) & 0.199 (0.00) & 0.132 (0.00) & \textcolor{red}{0.153 (0.00)}\tabularnewline
Pred-LLM & 0.041 (0.00) & 0.032 (0.00) & 0.061 (0.00) & 0.046 (0.00) & 0.043 (0.00) & 0.091 (0.00) & 0.038 (0.00) & \textcolor{red}{0.050 (0.00)}\tabularnewline
\rowcolor{even_color}SurvivalGAN & \uline{0.014} (0.00) & 0.034 (0.01) & 0.040 (0.00) & 0.063 (0.03) & 0.055 (0.00) & 0.092 (0.01) & 0.089 (0.02) & \textcolor{red}{0.055 (0.01)}\tabularnewline
Ashhad & \textbf{0.013} (0.00) & \textbf{0.016} (0.00) & \uline{0.025} (0.00) & \textbf{0.012} (0.00) & \textbf{0.020} (0.00) & 0.158 (0.00) & \uline{0.021} (0.00) & \textcolor{red}{\uline{0.038}}\textcolor{red}{{} (0.00)}\tabularnewline
\rowcolor{even_color}SurvDiff & 0.088 (0.01) & 0.037 (0.00) & 0.151 (0.01) & 0.123 (0.02) & \uline{0.041} (0.01) & 0.179 (0.04) & 0.024 (0.00) & \textcolor{red}{0.092 (0.01)}\tabularnewline
SurvTabGen & 0.026 (0.01) & \uline{0.023} (0.00) & \textbf{0.012} (0.00) & \uline{0.026} (0.03) & \textbf{0.020} (0.01) & \textbf{0.035} (0.02) & 0.040 (0.01) & \textbf{\textcolor{red}{0.026}}\textcolor{red}{{} (0.01)}\tabularnewline
\hline 
\end{tabular} }
\end{table}

\paragraph{Data quality}

\begin{figure}[th]
\begin{centering}
\includegraphics[width=0.8\textwidth]{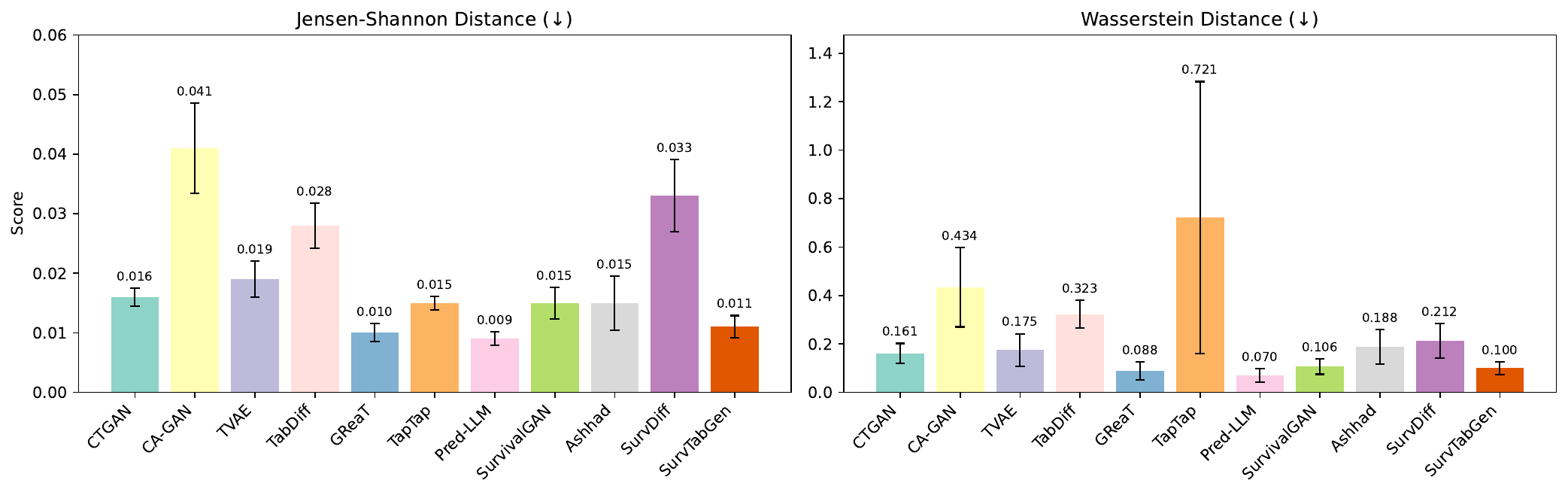}
\par\end{centering}
\caption{\label{fig:Data-quality-evaluation}Synthetic data quality measured
by the average Jensen-Shannon and Wasserstein distances across all
datasets. \textit{Lower values indicate higher distributional fidelity}.}
\end{figure}

As shown in Figure \ref{fig:Data-quality-evaluation}, SurvTabGen
outperforms all non-LLM methods. Two LLM-based methods GReaT and Pred-LLM
achieve slightly lower distributional distances. This is expected
since LLMs excel at capturing fine-grained tabular nuances, achieving
better distributional alignment at the cost of high computational
overhead. SurvTabGen prioritizes the preservation of multivariate
survival dynamics. While this introduces a minor trade-off in marginal
matching, the performance gap remains minimal ($\leq0.002$), and
is achieved at a fraction of the computational cost.

\paragraph{Privacy preservation}

Table \ref{tab:DCR-RIA} reports the average DCR and Re-identification
risk over all datasets. Compared to both SurvivalGAN and Ashhad, SurvTabGen
achieves a higher 5th percentile DCR and a lower Re-identification
risk, indicating a stronger defense against privacy leakage. While
SurvDiff registers the lowest Re-identification risk and the highest
DCR values, this must be interpreted cautiously. Its substantially
inflated DCR suggests that its synthetic samples have drifted far
from the real distribution, a conclusion supported by its poor data
quality metrics shown in Figure \ref{fig:Data-quality-evaluation}.
Overall, SurvTabGen achieves the most effective and meaningful balance
between privacy preservation and data fidelity among survival-oriented
methods.

\begin{table}[th]
\caption{\label{tab:DCR-RIA}Privacy evaluation using DCR and Re-identification
risk across all datasets.}

\centering{}\resizebox{\textwidth}{!}{{\renewcommand{\arraystretch}{1.2}%
\begin{tabular}{|l|c|c|c|c|c|c|c|c|c|c|c|}
\hline 
\multicolumn{1}{|l|}{} & \textbf{CTGAN} & \textbf{CA-GAN} & \textbf{TVAE} & \textbf{TabDiff} & \textbf{GReaT} & \textbf{TapTap} & \textbf{Pred-LLM} & \textbf{SurvivalGAN} & \textbf{Ashhad} & \textbf{SurvDiff} & \textbf{SurvTabGen}\tabularnewline
\hline 
DCR ($\uparrow$) & 0.105 (0.10) & 0.134 (0.09) & 0.052 (0.05) & 0.110 (0.08) & 0.059 (0.08) & 0.088 (0.06) & 0.051 (0.06) & 0.056 (0.05) & 0.067 (0.06) & 0.215 (0.16) & 0.128 (0.15)\tabularnewline
\hline 
\multicolumn{1}{|l|}{Risk ($\downarrow$)} & 0.198 (0.11) & 0.141 (0.08) & 0.251 (0.13) & 0.148 (0.10) & 0.359 (0.12) & 0.158 (0.16) & 0.409 (0.09) & 0.324 (0.16) & 0.331 (0.16) & 0.091 (0.08) & 0.307 (0.15)\tabularnewline
\hline 
\end{tabular}}}
\end{table}

\subsection{Ablation Studies}

\paragraph{Component analysis}

\begin{figure}[th]
\begin{centering}
\includegraphics[width=0.4\textwidth]{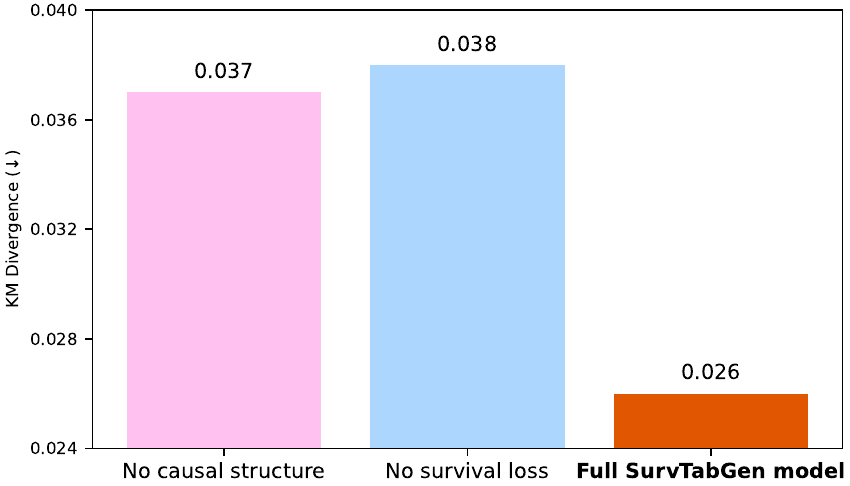}
\par\end{centering}
\caption{\label{fig:components}Impact of components on survival fidelity.
KM Divergence is averaged across all datasets.}

\end{figure}

We conduct an ablation to isolate the contribution of each component.
As shown in Figure \ref{fig:components}, removing causal discovery
module (\textbf{no causal structure}) degrades the survival fidelity,
highlighting the necessity of learning the underlying feature relationships.
Similarly, disabling the survival-aware objective (\textbf{no survival
loss} i.e., $\lambda=0$) leads to performance drops, confirming the
critical role of survival supervision in guiding temporal alignment.
The \textbf{full SurvTabGen model} achieves optimal performance (the
lowest KM Divergence), suggesting that both components are mutually
beneficial. \textit{(Note: Sensitivity analyses regarding the coefficient
$\lambda$ and alternative causal discovery algorithms are detailed
in Appendices \ref{subsec:Different-lambda} and \ref{subsec:Different-causal-discovery})}.

\paragraph{Visualization}

\begin{figure}[th]
\begin{centering}
\includegraphics[width=0.8\textwidth]{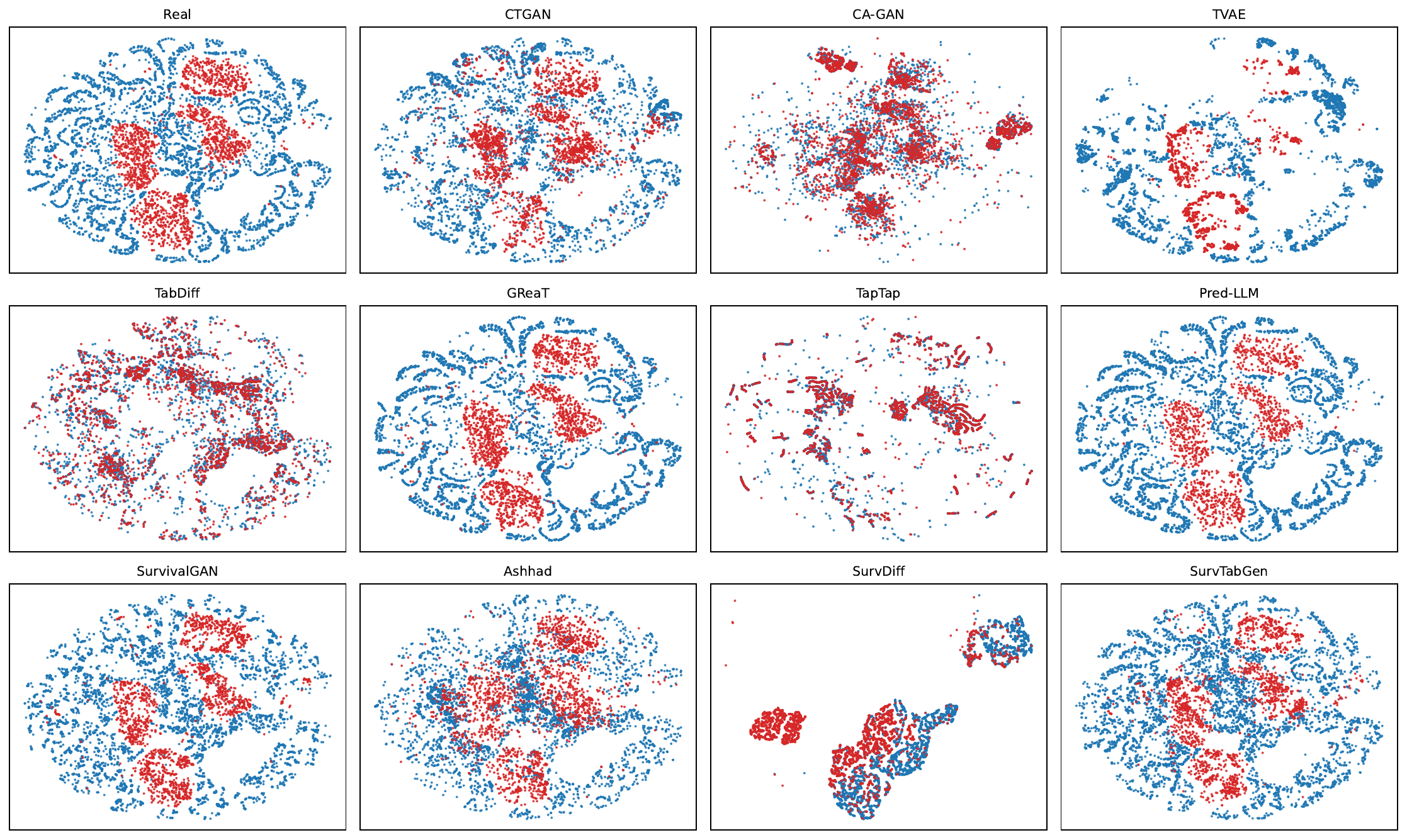}
\par\end{centering}
\caption{\label{fig:t-SNE-visualization}t-SNE visualization of real and synthetic
samples on the \textit{Flchain} dataset. Red indicates an observed
event ($E=1$) while blue indicates censoring ($E=0$). SurvTabGen
demonstrates superior coverage of both dense clusters and peripheral
data points.}
\end{figure}

Figure \ref{fig:t-SNE-visualization} presents a t-SNE visualization
\citep{VanderMaaten2008} of the synthetic samples generated by each
method on the \textit{Flchain} dataset. GReaT, Pred-LLM, SurvivalGAN,
and SurvTabGen successfully capture the major clusters observed in
the real data (\textit{first plot}), indicating a reasonable preservation
of dominant data structures. However, SurvivalGAN primarily reproduces
only the dense clustered regions, exhibiting weak coverage of boundary
areas and scattered, non-clustered samples. In contrast, SurvTabGen
preserves the primary clusters while consistently capturing the peripheral
and dispersed data points. This visual evidence further supports the
conclusion that SurvTabGen captures a more complete and representative
joint survival distribution than SurvivalGAN.

\subsection{Additional Experiments}

We report extra experiments in \textbf{Appendix}, including:
\begin{itemize}
\item \textbf{Other metrics:} \textit{Appendix} \ref{subsec:Survival-preservation}
reports \textit{Optimism} and \textit{Short-sightedness}.
\item \textbf{Subgroup-specific survival preservation:} \textit{Appendix}
\ref{subsec:Subgroup-Specific-Survival} reports performance at subpopulation
level.
\item \textbf{Survival estimators:} \textit{Appendix} \ref{subsec:Different-survival-estimator}
investigates the impact of alternative survival estimators.
\item \textbf{Robustness to causal graph ${\cal G}$:} \textit{Appendix}
\ref{subsec:Robustness-to-causal} examines the sensitivity to potential
errors in ${\cal G}$.
\item \textbf{Downstream utility:} \textit{Appendix} \ref{subsec:Downstream-performance}
reports downstream performance.
\item \textbf{Computational efficiency:} \textit{Appendix} \ref{subsec:Computational-efficiency}
compares the runtime of each method.
\end{itemize}

\section{Conclusion\label{sec:Conclusion}}

SurvTabGen provides an effective, end-to-end framework for synthetic
survival data generation. By unifying the generation of covariates,
event indicators, and survival times into a single joint process,
it overcomes the disjointed architectures of prior methods. SurvTabGen
embeds structural clinical reality via causal discovery while simultaneously
enforcing macroscopic temporal alignment through a reinforcement learning-driven
survival objective, minimizing Kaplan-Meier curve discrepancies via
policy gradients. Extensive evaluations against 10 SOTA baselines
across seven diverse real-world datasets show that SurvTabGen demonstrates
strong and consistent performance across diverse survival generation
benchmarks. It consistently achieves superior survival preservation
while maintaining competitive distributional quality and robust privacy
protection.

\bibliographystyle{plainnat}
\bibliography{references}

\appendix

\section{Proof for Proposition \ref{prop:wasserstein_iad}\label{sec:Proof-for-Theorem}}

For one-dimensional probability measures on $\mathbb{R}$ with finite
first moments, the 1-Wasserstein distance is the integral of the absolute
difference between their cumulative distribution functions (CDFs).
Let $F_{\mathrm{real}}(t)=1-S_{\mathrm{real}}(t)$ and $F_{\mathrm{syn}}(t)=1-S_{\mathrm{syn}}(t)$
be the respective CDFs. Thus:
\begin{align*}
\mathcal{W}_{1}(P_{T},Q_{T}) & =\int_{-\infty}^{\infty}\left|F_{\mathrm{real}}(t)-F_{\mathrm{syn}}(t)\right|dt
\end{align*}

As $T\geq0$, the lower bound of integration becomes zero:
\begin{align*}
\mathcal{W}_{1}(P_{T},Q_{T}) & =\int_{0}^{\infty}\left|F_{\mathrm{real}}(t)-F_{\mathrm{syn}}(t)\right|dt\\
 & =\int_{0}^{\infty}\left|(1-S_{\mathrm{real}}(t))-(1-S_{\mathrm{syn}}(t))\right|dt\\
 & =\int_{0}^{\infty}\left|S_{\mathrm{syn}}(t)-S_{\mathrm{real}}(t)\right|dt\\
 & =\mathrm{IAD}
\end{align*}

In practical survival settings, the survival functions $S_{\mathrm{real}}(t)$
and $S_{\mathrm{syn}}(t)$ are not observable and are instead estimated
via Kaplan-Meier estimators. Therefore, the resulting empirical IAD
provides a nonparametric approximation of the above quantity under
standard assumptions (e.g., independent censoring).

\section{Formulation of the Auxiliary Penalty $\mathcal{P}(\hat{\mathcal{B}})$\label{sec:aux_penalty}}

To stabilize the policy gradient, we define an auxiliary penalty $\mathcal{P}(\hat{\mathcal{B}})$.
This joint penalty minimizes discrepancies in the marginal event rates
and conditional time distributions between the generated batch $\hat{\mathcal{B}}$
and the real dataset $\mathcal{D}$. We set the generated batch size
equal to the real dataset size (i.e., $B=N$) to ensure stable, full-population
metric evaluations.

The penalty consists of two components. The \textbf{event penalty}
$\mathcal{P}_{E}$ measures the absolute difference in expected event
rates:
\begin{equation}
\mathcal{P}_{E}(\hat{\mathcal{B}})=\left|\frac{1}{N}\sum_{i=1}^{N}e_{i}-\frac{1}{N}\sum_{j=1}^{N}\hat{e}_{j}\right|
\end{equation}

The \textbf{time penalty} $\mathcal{P}_{T}$ aligns the temporal distributions
by minimizing the absolute difference across a set of $Q$ quantiles
of the survival times:
\begin{equation}
\mathcal{P}_{T}(\hat{\mathcal{B}})=\sum_{q\in Q}\left|\mathrm{Quantile}_{q}(T)-\mathrm{Quantile}_{q}(\hat{T})\right|
\end{equation}

The final joint auxiliary penalty is the sum of these components:
\begin{equation}
\mathcal{P}(\hat{\mathcal{B}})=\mathcal{P}_{E}(\hat{\mathcal{B}})+\mathcal{P}_{T}(\hat{\mathcal{B}})
\end{equation}

\section{Metrics\label{sec:Metrics}}

We summarize survival evaluation metrics used in this work. They are
standard metrics in survival analysis and synthetic survival data
evaluation \citep{norcliffe2023survivalgan,ashhad2025generating,brockschmidt2026survdiff}.

\subsection{Kaplan-Meier (KM) divergence}

KM divergence measures the discrepancy between survival curves estimated
from real and synthetic data:
\begin{equation}
\text{KM Divergence}=\frac{1}{T}\int_{0}^{T}\left|S_{\text{syn}}(t)-S_{\text{real}}(t)\right|dt,\label{eq:KM_divergence}
\end{equation}
where $S_{\text{real}}(t)$ and $S_{\text{syn}}(t)$ are Kaplan-Meier
survival functions, and $T$ is the maximum observed time horizon.
\textit{Lower values indicate better alignment of survival dynamics}.

\subsection{Optimism}

Optimism captures bias in survival estimation:
\begin{equation}
\text{Optimism}=\frac{1}{T}\int_{0}^{T}\left(S_{\text{syn}}(t)-S_{\text{real}}(t)\right)dt\label{eq:Optimism}
\end{equation}

Positive values indicate \textit{over-optimism} (synthetic survival
is higher) while negative values indicate \textit{over-pessimism}.
Values closer to 0 are preferred.

\subsection{Short-Sightedness}

Short-Sightedness measures how well the synthetic data covers the
temporal range of real data:
\begin{equation}
\text{Short-Sightedness}=\frac{T_{\text{real}}-T_{\text{syn}}}{T_{\text{real}}},\label{eq:sightedness}
\end{equation}
where $T_{\text{real}}$ is the maximum observed time in the real
dataset and $T_{\text{syn}}$ is the corresponding maximum time in
the synthetic dataset. Values close to 0 indicate good coverage while
larger values suggest early truncation of events.

\section{Data Description \label{sec:Data-descriptions}}

We use seven real-world survival datasets with diverse characteristics
in terms of size, censoring ratio, and feature composition, covering
both clinical and industrial domains. They are summarized in Table
\ref{tab:datasets}.

Overall, the datasets vary significantly in scale (from 100 to 7,874
instances), degree of censoring (from low to highly imbalanced), and
feature types (mixed continuous-categorical or fully continuous).
This diversity ensures a comprehensive evaluation setting, allowing
us to assess the robustness and generalization ability of the proposed
method across different survival data scenarios.

\begin{table}[th]
\caption{\label{tab:datasets}Summary statistics of the datasets used in the
study.}

\centering{}\resizebox{\textwidth}{!}{{\renewcommand{\arraystretch}{1.2}%
\begin{tabular}{|l|r|r|r|r|r|}
\hline 
Dataset & No. instances & No. censored instances & No. features & No. continuous features & No. categorical features\tabularnewline
\hline 
AIDS & 1,151 & 1,055 & 11 & 3 & 8\tabularnewline
Cgvhd & 100 & 4 & 4 & 1 & 3\tabularnewline
Flchain & 7,874 & 5,705 & 9 & 4 & 5\tabularnewline
Gbsg2 & 686 & 387 & 8 & 5 & 3\tabularnewline
Veterans & 137 & 9 & 6 & 3 & 3\tabularnewline
Whas500 & 500 & 285 & 14 & 6 & 8\tabularnewline
AM & 3,245 & 2,780 & 10 & 10 & 0\tabularnewline
\hline 
\end{tabular}}}
\end{table}

\section{Additional Experiments}

\subsection{Survival preservation\label{subsec:Survival-preservation}}

In addition to the KM Divergence metric reported in the main paper,
we further evaluate survival preservation using Optimism and Short-sightedness
metrics (described in \textit{Appendix} \ref{sec:Metrics}), where
values closer to zero indicate better alignment with real survival
behavior.

As shown in Table \ref{tab:Optimism-Sighted}, SurvTabGen achieves
the average values closest to zero on both metrics, indicating balanced
and unbiased survival modeling. Other methods (especially SurvivalGAN)
exhibit larger deviations, reflecting either over-optimistic or over-pessimistic
behavior. These results further support the effectiveness of SurvTabGen
in preserving realistic survival dynamics.

\begin{table}[th]
\caption{\label{tab:Optimism-Sighted}Optimism and Short-Sightedness metrics
showing the survival preservation performance of each method across
datasets.}

\centering{}\resizebox{\textwidth}{!}{{\renewcommand{\arraystretch}{1.2}%
\begin{tabular}{|c|l|r|r|r|r|r|r|r|r|}
\hline 
Metrics & Methods & AIDS & Cgvhd & Flchain & Gbsg2 & Veterans & Whas500 & AM & \textcolor{red}{AVG}\tabularnewline
\hline 
\multirow{11}{*}{\begin{cellvarwidth}[t]
\centering
Optimism

(\textrightarrow{} 0)
\end{cellvarwidth}} & CTGAN & \uline{-0.019} (0.02) & 0.020 (0.07) & -0.043 (0.02) & -0.040 (0.07) & 0.064 (0.10) & -0.094 (0.06) & -0.104 (0.01) & \textcolor{red}{-0.031 (0.05)}\tabularnewline
 & CA-GAN & -0.129 (0.00) & 0.263 (0.19) & -0.076 (0.01) & -0.089 (0.14) & 0.300 (0.08) & \uline{-0.006} (0.11) & -0.062 (0.01) & \textcolor{red}{0.029 (0.01)}\tabularnewline
 & TVAE & 0.055 (0.01) & -0.051 (0.00) & -0.056 (0.01) & 0.054 (0.02) & 0.076 (0.00) & 0.032 (0.02) & 0.053 (0.02) & \textcolor{red}{0.023 (0.01)}\tabularnewline
 & TabDiff & -0.146 (0.07) & 0.114 (0.13) & \textbf{0.008} (0.04) & 0.085 (0.23) & 0.262 (0.10) & 0.044 (0.14) & -0.283 (0.06) & \textbf{\textcolor{red}{0.012}}\textcolor{red}{{} (0.11)}\tabularnewline
 & GReat & 0.022 (0.00) & -0.033 (0.00) & 0.022 (0.00) & -0.073 (0.00) & -0.043 (0.00) & -0.090 (0.00) & \textbf{-0.008} (0.00) & \textcolor{red}{-0.029 (0.00)}\tabularnewline
 & TapTap & -0.065 (0.00) & -0.142 (0.00) & -0.174 (0.00) & -0.257 (0.00) & -0.100 (0.00) & -0.199 (0.00) & 0.033 (0.00) & \textcolor{red}{-0.129 (0.00)}\tabularnewline
 & Pred-LLM & 0.040 (0.00) & 0.600 (0.00) & 0.020 (0.00) & 0.080 (0.00) & 0.590 (0.00) & 0.070 (0.00) & -0.034 (0.00) & \textcolor{red}{0.195 (0.00)}\tabularnewline
 & SurvivalGAN & \textbf{-0.009} (0.01) & -0.025 (0.01) & -0.035 (0.00) & -0.037 (0.04) & -0.045 (0.01) & -0.031 (0.02) & -0.089 (0.02) & \textcolor{red}{-0.039 (0.02)}\tabularnewline
 & Ashhad & \textbf{-0.009} (0.00) & \uline{-0.012} (0.00) & -0.012 (0.00) & \textbf{0.008} (0.00) & \textbf{-0.004} (0.00) & 0.158 (0.00) & \textbf{-0.008} (0.00) & \textcolor{red}{\uline{0.017}}\textcolor{red}{{} (0.00)}\tabularnewline
 & SurvDiff & -0.086 (0.01) & -0.032 (0.01) & -0.145 (0.01) & -0.123 (0.02) & -0.035 (0.01) & -0.009 (0.07) & \uline{-0.016} (0.00) & \textcolor{red}{-0.064 (0.02)}\tabularnewline
 & SurvTabGen & \uline{-0.019} (0.01) & \textbf{-0.009} (0.01) & \uline{-0.010} (0.01) & \uline{-0.017} (0.02) & \uline{0.006} (0.01) & \textbf{0.002} (0.03) & -0.037 (0.02) & \textbf{\textcolor{red}{-0.012}}\textcolor{red}{{} (0.01)}\tabularnewline
\hline 
\multirow{11}{*}{\begin{cellvarwidth}[t]
\centering
Short-Sightedness

(\textrightarrow{} 0)
\end{cellvarwidth}} & CTGAN & \textbf{0.000} (0.00) & 0.238 (0.12) & 0.002 (0.00) & \textbf{0.000} (0.00) & 0.156 (0.22) & \textbf{0.002} (0.00) & \uline{0.002} (0.00) & \textcolor{red}{0.057 (0.05)}\tabularnewline
 & CA-GAN & 0.142 (0.01) & 0.364 (0.09) & 0.038 (0.00) & 0.294 (0.02) & 0.310 (0.12) & 0.110 (0.03) & 0.072 (0.03) & \textcolor{red}{0.190 (0.04)}\tabularnewline
 & TVAE & \textbf{0.000} (0.00) & 0.808 (0.01) & 0.006 (0.01) & \textbf{0.000} (0.00) & 0.820 (0.03) & 0.024 (0.02) & \textbf{0.000} (0.00) & \textcolor{red}{0.237 (0.01)}\tabularnewline
 & TabDiff & \uline{0.002} (0.00) & 0.140 (0.17) & \textbf{0.000} (0.00) & \uline{0.004} (0.01) & 0.148 (0.15) & \textbf{0.002} (0.00) & 0.016 (0.02) & \textcolor{red}{0.045 (0.05)}\tabularnewline
 & GReat & 0.040 (0.00) & 0.330 (0.00) & 0.010 (0.00) & 0.110 (0.00) & 0.300 (0.00) & 0.080 (0.00) & 0.020 (0.00) & \textcolor{red}{0.127 (0.00)}\tabularnewline
 & TapTap & 0.030 (0.00) & 0.850 (0.00) & 0.010 (0.00) & 0.180 (0.00) & 0.750 (0.00) & 0.060 (0.00) & \textbf{0.000} (0.00) & \textcolor{red}{0.269 (0.00)}\tabularnewline
 & Pred-LLM & 0.041 (0.00) & \textbf{-0.024} (0.00) & 0.061 (0.00) & -0.045 (0.00) & \textbf{-0.043} (0.00) & \uline{0.016} (0.00) & 0.020 (0.00) & \textbf{\textcolor{red}{0.004}}\textcolor{red}{{} (0.00)}\tabularnewline
 & SurvivalGAN & 0.006 (0.00) & 0.521 (0.10) & \uline{0.001} (0.00) & 0.052 (0.02) & 0.550 (0.10) & 0.024 (0.02) & 0.012 (0.01) & \textcolor{red}{0.167 (0.04)}\tabularnewline
 & Ashhad & -0.109 (0.00) & 0.295 (0.00) & -0.041 (0.00) & -0.114 (0.00) & \uline{0.051} (0.00) & -0.218 (0.00) & -0.038 (0.00) & \textcolor{red}{\uline{-0.025}}\textcolor{red}{{} (0.00)}\tabularnewline
 & SurvDiff & 0.153 (0.02) & 0.514 (0.10) & 0.335 (0.01) & 0.080 (0.05) & 0.409 (0.26) & -0.310 (0.30) & -0.010 (0.00) & \textcolor{red}{0.167 (0.11)}\tabularnewline
 & SurvTabGen & 0.008 (0.03) & \uline{0.042} (0.23) & 0.004 (0.01) & 0.007 (0.05) & -0.075 (0.16) & -0.033 (0.05) & 0.016 (0.02) & \textbf{\textcolor{red}{-0.004}}\textcolor{red}{{} (0.08)}\tabularnewline
\hline 
\end{tabular}}}
\end{table}

\subsection{Survival preservation on subgroups\label{subsec:Subgroup-Specific-Survival}}

Since preserving survival patterns for subpopulations is important
in some applications, we demonstrate SurvTabGen's performance at subpopulation
level. 

On the Whas500 dataset, we evaluate three age cohorts: \textit{Young}
($<65$), \textit{Elderly} ($65$--$74$), and \textit{Very Elderly}
($\geq75$). As shown in Table \ref{tab:Subgroup-specific-KM-Divergence},
SurvTabGen accurately preserves subgroup-specific survival distributions,
achieving the lowest KM Divergence on the overall population, as well
as on the Elderly and Very Elderly subgroups. These results suggest
that optimizing the global survival objective also helps preserve
subgroup-specific survival patterns.

\begin{table}[th]
\caption{\label{tab:Subgroup-specific-KM-Divergence}Subgroup-specific KM Divergence
($\downarrow$) on the Whas500 dataset across age cohorts.}

\centering{}\resizebox{0.8\textwidth}{!}{%
\begin{tabular}{|l|c|c|c|c|}
\hline 
\rowcolor{header_color} & Ashhad & SurvivalGAN & SurvDiff & SurvTabGen\tabularnewline
\hline 
All Ages & 0.158 (0.00) & 0.092 (0.01) & 0.179 (0.04) & \textbf{0.035 (0.02)}\tabularnewline
\hline 
Young ($<65$) & 0.118 (0.01) & \textbf{0.063 (0.02)} & 0.283 (0.07) & 0.131 (0.02)\tabularnewline
\hline 
Elderly ($65$--$74$) & 0.077 (0.00) & 0.121 (0.04) & 0.180 (0.05) & \textbf{0.076 (0.04)}\tabularnewline
\hline 
Very Elderly ($\geq75$) & 0.239 (0.00) & 0.131 (0.04) & 0.121 (0.02) & \textbf{0.079 (0.01)}\tabularnewline
\hline 
\end{tabular}}
\end{table}

\subsection{Different values for $\lambda$\label{subsec:Different-lambda}}

\begin{figure}[th]
\begin{centering}
\includegraphics[width=0.6\textwidth]{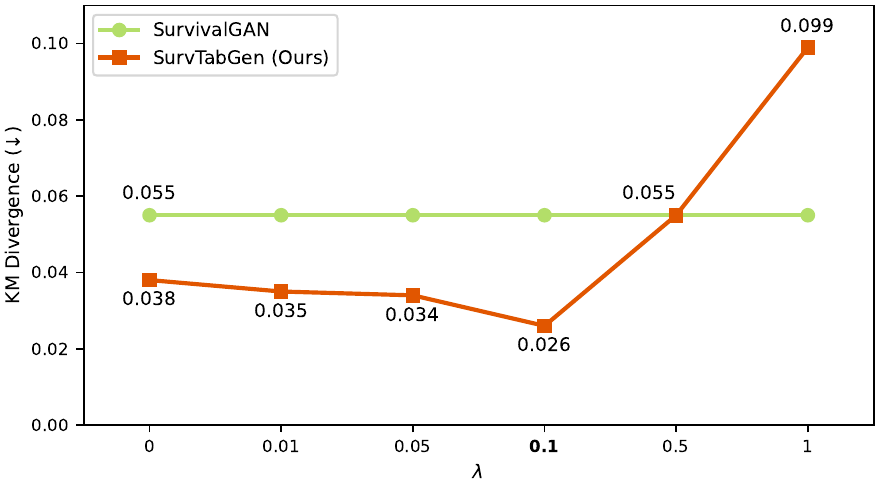}
\par\end{centering}
\caption{\label{fig:Lambda-KM}Influence of $\lambda$ on SurvTabGen's performance.
The average KM Divergence is computed across all datasets.}
\end{figure}

We conduct an ablation study on the survival loss weight $\lambda$,
which balances the survival loss with the adversarial loss. As shown
in Figure \ref{fig:Lambda-KM}, increasing $\lambda$ initially improves
performance, with KM divergence decreasing from 0.038 to 0.026 as
$\lambda$ increases to 0.1, indicating better alignment of temporal
dynamics. Importantly, even at $\lambda=0$, SurvTabGen already outperforms
SurvivalGAN, suggesting that generation based on a structured causal
process is more effective than multi-stage conditional generation.
However, further increasing $\lambda$ degrades performance as the
model overemphasizes the survival objective. Overall, SurvTabGen remains
stably in a large range of $\lambda\in[0,0.1]$, where it is always
better than SurvivalGAN.

\subsection{Different causal discovery methods\label{subsec:Different-causal-discovery}}

Besides the PC algorithm \citep{spirtes2000causation} used in our
framework, we explore alternative causal discovery methods such as
Greedy Equivalence Search (GES) \citep{Andrews2023}, Fast Causal
Inference (FCI) \citep{Spirtes2013}, and Best Order Score Search
(BOSS) \citep{Chickering2002}. As shown in Table \ref{tab:Causal-discovery},
PC achieves the lowest average KM divergence but other methods also
perform well. Overall, SurvTabGen is robust to different causal discovery
methods.

\begin{table}[th]
\caption{\label{tab:Causal-discovery}Impact of causal discovery method on
SurvTabGen's performance. The average KM Divergence is computed across
all datasets.}

\centering{}\resizebox{0.8\textwidth}{!}{%
\begin{tabular}{|c|l|r|r|r|r|r|r|r|r|}
\hline 
Metric & Method & AIDS & Cgvhd & Flchain & Gbsg2 & Veterans & Whas500 & AM & \textcolor{red}{AVG}\tabularnewline
\hline 
\hline 
\multirow{4}{*}{\begin{cellvarwidth}[t]
\centering
KM Divergence

($\downarrow$)
\end{cellvarwidth}} & BOSS & \uline{0.024} & 0.030 & 0.043 & 0.031 & \uline{0.027} & \uline{0.041} & \uline{0.039} & \textcolor{red}{0.033}\tabularnewline
 & FCI & \textbf{0.023} & 0.031 & 0.075 & 0.030 & 0.034 & 0.048 & \textbf{0.031} & \textcolor{red}{0.039}\tabularnewline
 & GES & 0.029 & \uline{0.028} & \uline{0.031} & \uline{0.028} & 0.027 & \textbf{0.038} & 0.041 & \textcolor{red}{\uline{0.032}}\tabularnewline
 & PC & 0.026 & \textbf{0.023} & \textbf{0.016} & \textbf{0.026} & \textbf{0.020} & 0.043 & 0.040 & \textbf{\textcolor{red}{0.026}}\tabularnewline
\hline 
\end{tabular}}
\end{table}

\subsection{Different survival estimators\label{subsec:Different-survival-estimator}}

\begin{figure}[th]
\begin{centering}
\includegraphics[width=0.6\textwidth]{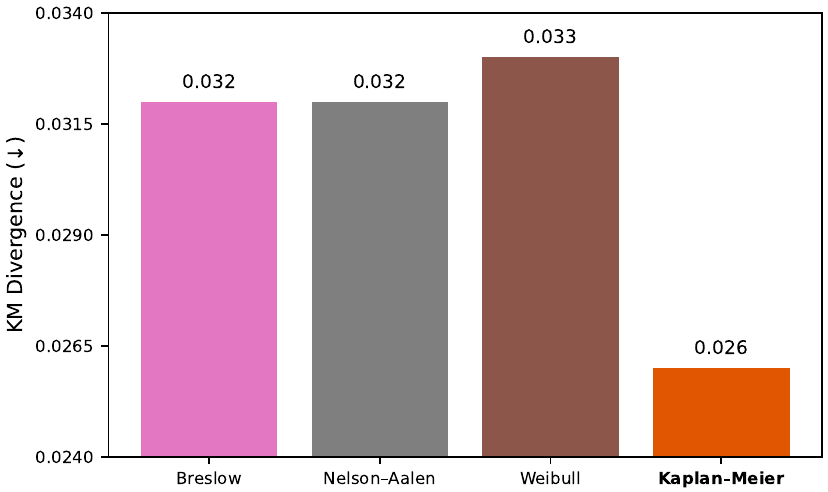}
\par\end{centering}
\caption{\label{fig:Survival-estimator}Influence of survival estimator choice
on SurvTabGen's performance. The average KM Divergence is computed
across all datasets.}
\end{figure}

We replace the default Kaplan-Meier estimator with alternative survival
estimators, including Breslow \citep{fleming2013counting}, Nelson-Aalen
\citep{aalen2008survival}, and Weibull \citep{kalbfleisch2002statistical}
estimators. Any of these estimators can be used to compute the survival
function, which defines the reward. As shown in Figure \ref{fig:Survival-estimator},
the Kaplan-Meier estimator reaches the lowest KM Divergence while
other estimators lead to degraded performance. This highlights the
advantage of the Kaplan-Meier estimator as a non-parametric estimator
that does not assume a fixed distribution and can accurately handle
censored data.

\subsection{Robustness to causal graph\label{subsec:Robustness-to-causal}}

To evaluate the robustness of SurvTabGen to causal graph quality,
we construct a synthetic dataset (10K samples), which consists of
a covariate group $X=\{A,C,F\}$, one event indicator $E$, and one
survival time variable $T$. The internal relationships among $A$,
$C$, and $F$ are kept fixed to preserve the structure of the covariate
space. We then perturb the causal dependencies among $X$, $E$, and
$T$ while maintaining the ground-truth survival mechanism $X\rightarrow E$
and $(X,E)\rightarrow T$.

Table \ref{tab:Graph-quality-synthetic} shows that multiple graph
variants are generated by progressively degrading the causal structure
(e.g., reversing directions, removing edges, or introducing incorrect
chains) without altering the underlying data generation process within
$X$.

\begin{table}[th]
\caption{\label{tab:Graph-quality-synthetic}Graph-quality spectrum for a five-variable
causal graph derived from the synthetic dataset. A, C, F, E, and T
denotes Age, Comorbidity, Frailty, Event, and Time, respectively.}

\centering{}\resizebox{\textwidth}{!}{{\renewcommand{\arraystretch}{1.2}%
\begin{tabular}{|l|c|c|c|c|c|c|}
\hline 
\textbf{Accuracy level} & \textbf{100\%} & \textbf{80\%} & \textbf{60\%} & \textbf{40\%} & \textbf{20\%} & \textbf{0\%}\tabularnewline
\hline 
\hline 
Graph Structure & \includegraphics{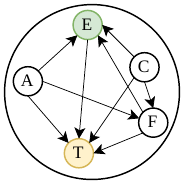} & \includegraphics{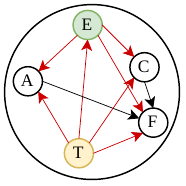} & \includegraphics{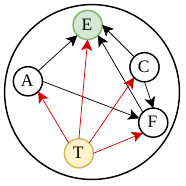} & \includegraphics{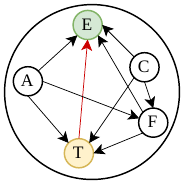} & \includegraphics{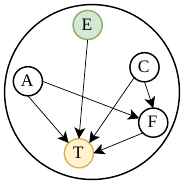} & \includegraphics{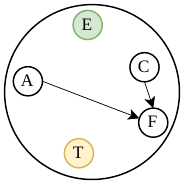}\tabularnewline
\hline 
Description & Ground truth - true mechanism & Predicting causes from effect & Effect treated as cause & One wrong chain & Missed one parent & No edge found\tabularnewline
\hline 
\end{tabular}}}
\end{table}

\begin{figure}[th]
\begin{centering}
\includegraphics[width=0.6\textwidth]{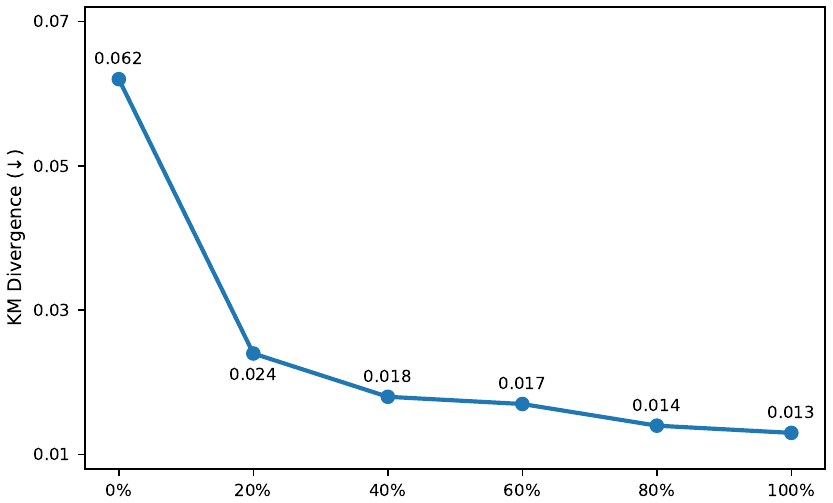}
\par\end{centering}
\caption{\label{fig:Robustness-evaluation-synthetic}SurvTabGen's performance
on a synthetic dataset under perturbed causal structures.}
\end{figure}

We evaluate the impact of these perturbations in Figure \ref{fig:Robustness-evaluation-synthetic}.
It shows that when the causal graph is closer to the ground truth
(i.e., \textit{graph-accuracy level approaches 100\%}), KM Divergence
is significantly reduced. In contrast, distorted graph structures
lead to degradation in KM Divergence. However, SurvTabGen's performance
remains highly robust across a wide spectrum of graph-accuracy levels,
even when the graph accuracy drops to 20\%.

These results suggest that accurate modeling of the dependencies between
$X$, $E$, and $T$ is critical for preserving survival dynamics.
More importantly, it also demonstrates the robustness of SurvTabGen
under varying causal graph quality.

\subsection{Downstream performance\label{subsec:Downstream-performance}}

We train three models Cox Proportional Hazards (CoxPH) \citep{cox1972regression},
SurvivalXGBoost \citep{barnwal2022survival}, and Random Survival
Forest \citep{ishwaran2008random} on the synthetic data and evaluate
them on real held-out data. To align with SurvivalGAN, Ashhad, and
SurvDiff evaluation protocols, we use a 67/33 data split (i.e., 2-fold
for training and 1-fold for testing) across all datasets, except for
the small dataset Cgvhd, where we use a 75/25 split to ensure sufficient
training samples. The hyperparameters for these survival models are
set exactly as detailed in SurvivalGAN and Ashhad. Performance is
measured using \textit{C-index} ($\uparrow$) \citep{harrell1982evaluating}
and \textit{Brier Score} ($\downarrow$) \citep{glenn1950verification},
reporting the best performance achieved across the three models.

As shown in Table \ref{tab:Brier}, SurvTabGen consistently achieves
strong performance under both metrics, remaining competitive with
SurvDiff in C-index and TapTap in Brier Score.

\begin{table}[th]
\caption{\label{tab:Brier}C-index and Brier Score metrics showing downstream
utility performance of each method across datasets. ``\textit{Real
Data}'' indicates the survival models are trained directly on real
data.}

\centering{}\resizebox{\textwidth}{!}{{\renewcommand{\arraystretch}{1.2}%
\begin{tabular}{|c|l|r|r|r|r|r|r|r|r|}
\hline 
Metrics & Methods & AIDS & Cgvhd & Flchain & Gbsg2 & Veterans & Whas500 & AM & \textcolor{red}{AVG}\tabularnewline
\hline 
\multirow{12}{*}{\begin{cellvarwidth}[t]
\centering
C-index

(\textuparrow )
\end{cellvarwidth}} & \textit{Real Data} & \textit{0.738 (0.04)} & \textit{0.500 (0.00)} & \textit{0.934 (0.00)} & \textit{0.672 (0.02)} & \textit{0.700 (0.03)} & \textit{0.769 (0.01)} & \textit{0.794 (0.03)} & \textit{\textcolor{red}{0.729}}\tabularnewline
\cline{2-10}
 & CTGAN & 0.567 (0.08) & 0.549 (0.09) & 0.929 (0.00) & 0.498 (0.11) & 0.524 (0.09) & 0.633 (0.06) & 0.590 (0.10) & \textcolor{red}{0.613}\tabularnewline
 & CA-GAN & 0.550 (0.11) & 0.501 (0.11) & 0.758 (0.01) & 0.516 (0.11) & 0.525 (0.06) & 0.483 (0.11) & 0.635 (0.09) & \textcolor{red}{0.567}\tabularnewline
 & TVAE & 0.697 (0.03) & 0.525 (0.07) & 0.927 (0.00) & 0.647 (0.02) & 0.613 (0.06) & \textbf{0.746} (0.03) & 0.722 (0.02) & \textcolor{red}{0.697}\tabularnewline
 & TabDiff & 0.479 (0.13) & 0.513 (0.05) & 0.547 (0.20) & 0.572 (0.08) & 0.546 (0.10) & 0.562 (0.07) & 0.484 (0.05) & \textcolor{red}{0.529}\tabularnewline
 & GReaT & 0.444 (0.07) & 0.500 (0.00) & 0.929 (0.00) & 0.549 (0.05) & 0.581 (0.04) & 0.642 (0.07) & 0.634 (0.06) & \textcolor{red}{0.611}\tabularnewline
 & TapTap & 0.608 (0.12) & 0.523 (0.05) & 0.930 (0.00) & 0.571 (0.06) & 0.540 (0.05) & 0.642 (0.07) & 0.688 (0.02) & \textcolor{red}{0.643}\tabularnewline
 & Pred-LLM & 0.674 (0.05) & \textbf{0.584} (0.12) & \uline{0.932} (0.00) & 0.580 (0.03) & 0.500 (0.00) & 0.639 (0.03) & \textbf{0.747} (0.02) & \textcolor{red}{0.665}\tabularnewline
 & SurvivalGAN & 0.631 (0.02) & 0.501 (0.09) & 0.928 (0.00) & 0.652 (0.03) & \textbf{0.656} (0.06) & 0.734 (0.02) & 0.673 (0.08) & \textcolor{red}{0.682}\tabularnewline
 & Ashhad & 0.532 (0.09) & 0.500 (0.00) & 0.894 (0.03) & 0.442 (0.03) & 0.541 (0.08) & 0.571 (0.09) & 0.604 (0.17) & \textcolor{red}{0.583}\tabularnewline
 & SurvDiff & \textbf{0.710} (0.03) & 0.500 (0.00) & \textbf{0.937} (0.00) & \textbf{0.665} (0.02) & \uline{0.655} (0.08) & \uline{0.742} (0.03) & \uline{0.742} (0.03) & \textbf{\textcolor{red}{0.711}}\tabularnewline
 & SurvTabGen & \uline{0.707} (0.05) & \uline{0.551} (0.09) & 0.929 (0.00) & \uline{0.656} (0.02) & 0.641 (0.03) & 0.728 (0.01) & 0.711 (0.05) & \textcolor{red}{\uline{0.703}}\tabularnewline
\hline 
\multirow{12}{*}{\begin{cellvarwidth}[t]
\centering
Brier Score

(\textdownarrow )
\end{cellvarwidth}} & \textit{Real Data} & \textit{0.060 (0.00)} & \textit{0.519 (0.21)} & \textit{0.102 (0.00)} & \textit{0.271 (0.01)} & \textit{0.116 (0.02)} & \textit{0.179 (0.01)} & \textit{0.058 (0.00)} & \textit{\textcolor{red}{0.186}}\tabularnewline
\cline{2-10}
 & CTGAN & 0.065 (0.00) & 0.168 (0.03) & 0.054 (0.00) & 0.267 (0.02) & 0.203 (0.08) & 0.244 (0.02) & 0.169 (0.01) & \textcolor{red}{0.167}\tabularnewline
 & CA-GAN & \textbf{0.058} (0.00) & 0.296 (0.07) & 0.124 (0.01) & 0.193 (0.01) & 0.345 (0.12) & 0.265 (0.03) & 0.099 (0.01) & \textcolor{red}{0.197}\tabularnewline
 & TVAE & 0.067 (0.00) & 0.152 (0.03) & 0.099 (0.00) & \uline{0.183} (0.01) & 0.231 (0.10) & \textbf{0.187} (0.01) & \uline{0.068} (0.01) & \textcolor{red}{0.141}\tabularnewline
 & TabDiff & 0.066 (0.00) & 0.158 (0.06) & 0.142 (0.00) & 0.202 (0.02) & 0.221 (0.12) & 0.287 (0.01) & 0.149 (0.02) & \textcolor{red}{0.175}\tabularnewline
 & GReaT & 0.067 (0.00) & 0.521 (0.21) & \textbf{0.048} (0.00) & 0.199 (0.01) & 0.144 (0.02) & 0.230 (0.02) & 0.080 (0.01) & \textcolor{red}{0.184}\tabularnewline
 & TapTap & 0.064 (0.00) & \uline{0.122} (0.01) & \textbf{0.048} (0.00) & 0.195 (0.01) & 0.151 (0.04) & 0.219 (0.00) & 0.080 (0.00) & \textbf{\textcolor{red}{0.126}}\tabularnewline
 & Pred-LLM & 0.066 (0.00) & \textbf{0.120} (0.01) & 0.100 (0.00) & 0.262 (0.00) & 0.625 (0.03) & 0.217 (0.01) & 0.071 (0.01) & \textcolor{red}{0.209}\tabularnewline
 & SurvivalGAN & 0.067 (0.01) & 0.160 (0.07) & \uline{0.049} (0.00) & \textbf{0.178} (0.00) & 0.191 (0.05) & 0.198 (0.01) & 0.181 (0.02) & \textcolor{red}{0.146}\tabularnewline
 & Ashhad & 0.066 (0.00) & 0.525 (0.20) & 0.105 (0.01) & 0.197 (0.01) & \uline{0.143} (0.03) & 0.236 (0.00) & 0.087 (0.01) & \textcolor{red}{0.194}\tabularnewline
 & SurvDiff & 0.066 (0.01) & 0.525 (0.20) & 0.108 (0.00) & 0.270 (0.01) & \textbf{0.135} (0.03) & \uline{0.189} (0.01) & \textbf{0.063} (0.01) & \textcolor{red}{0.194}\tabularnewline
 & SurvTabGen & \uline{0.061} (0.00) & 0.129 (0.02) & 0.050 (0.00) & 0.272 (0.01) & 0.144 (0.03) & 0.199 (0.01) & 0.083 (0.01) & \textcolor{red}{\uline{0.134}}\tabularnewline
\hline 
\end{tabular}}}
\end{table}

\subsection{Computational efficiency\label{subsec:Computational-efficiency}}

\begin{figure}[H]
\begin{centering}
\includegraphics[width=1\textwidth]{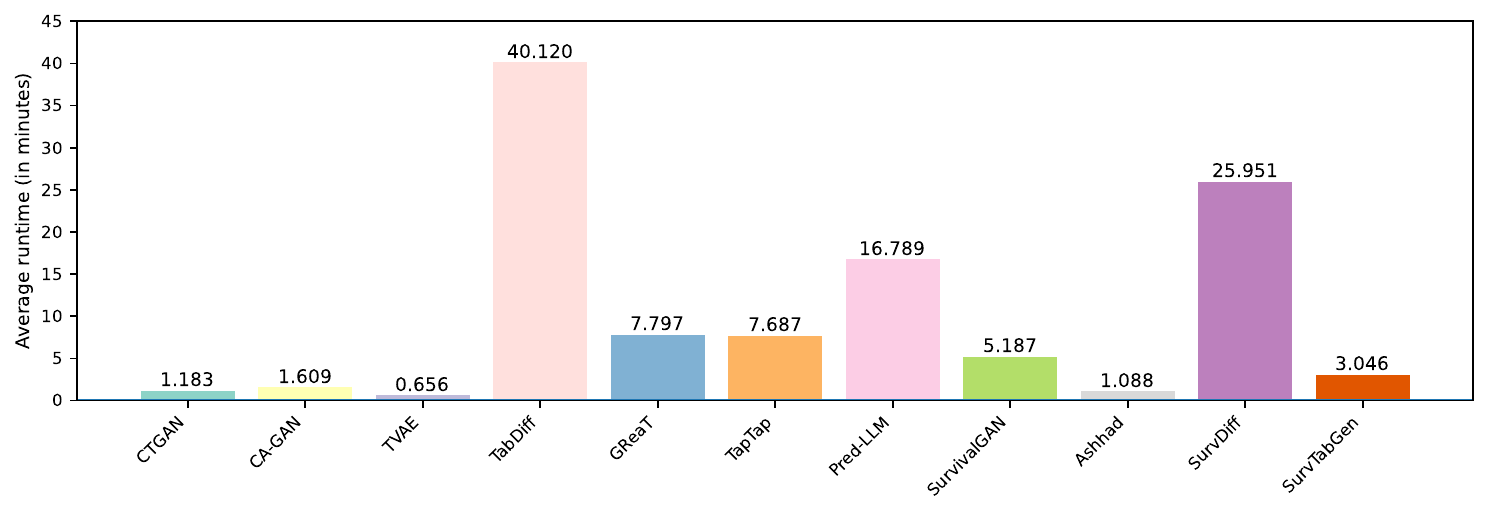}
\par\end{centering}
\caption{\label{fig:runtime}Average runtime (\textit{in minutes}) of each
method across all datasets.}
\end{figure}

Figure \ref{fig:runtime} presents the average wall-clock runtime
(\textit{in minutes}) of each method across seven datasets. All runtime
measurements are conducted on a machine with an NVIDIA Tesla V100
GPU, 8 CPU cores, and 64GB RAM. SurvTabGen is much faster than SurvivalGAN
(\textbf{1.7\texttimes{} faster}) and SurvDiff (\textbf{8.6\texttimes{}
faster}).

\end{document}